# Emergent Collaborative Deliberation in Multi-Model AI Systems

*A BFT-Derived Protocol for Human-Moderated Epistemic Synthesis*

**VD Doske** Independent Researcher / Consilia

---

## Abstract

We present the Consilium Protocol, a Byzantine Fault Tolerance-derived architecture for structured multi-model AI deliberation that treats inter-model disagreement as epistemic signal rather than error. The protocol assigns engineered cognitive personas to language models — separating *what a model is* from *how it reasons* — and introduces an In-Sample/Out-of-Sample validation framework adapted from quantitative finance to distinguish training-data consensus from empirically grounded conclusions. Across 1,478 deliberation sessions spanning 32 topics in 10 domain categories, we demonstrate that (1) the cognitive persona, not the underlying model, determines epistemic behavior: free edge-inference models costing $0.0002 per batch produced comparable analytical output to frontier models costing $10.69; (2) RLHF alignment training creates measurable, domain-specific epistemic blind spots — contested policy topics exhibit 12.3 percentage points less adversarial challenge than settled science topics, and AI safety topics show asymmetric bias (Δ=11.6%) where models challenge claims that AI is dangerous far more vigorously than claims that AI risk is overstated; (3) the protocol exhibits no directional bias of its own (immigration Δ=2.3%, renewables Δ=1.2%); and (4) out-of-sample evidence retrieval validated 239 claims with 100% evidence retrieval and surfaced 167 blind-spot discoveries invisible to training-data deliberation. Run-to-run reproducibility across randomized model×persona assignments averages ±2.2% standard deviation. Total cost for the complete battery including all overhead: $217. We release the protocol specification under MIT license to enable independent verification.

---

# 1. Introduction

What is truth?

The question is older than empiricism, older than the scientific method, older than the printing press that first democratized the ability to challenge established answers. For most of recorded history, the practical answer was institutional: truth was what the church declared, what the sovereign permitted, what the academy endorsed. The Enlightenment replaced institutional authority with empirical method — truth became what survived observation, replication, and open contestation. Eight decades of post-war liberal order cemented this bargain: free inquiry, adversarial debate, and the marketplace of ideas would, in aggregate, surface reliable knowledge.

That bargain is now under structural pressure from an unexpected direction. Not from censorship, propaganda, or institutional capture — though those persist — but from the architecture of the systems we built to accelerate knowledge production itself. Large language models trained on overlapping corpora produce outputs with uniform confidence regardless of whether the underlying knowledge is well-attested or hallucinated. A model that can render Shakespeare with the fluency of a scholar will render speculative geopolitics with identical conviction. It cannot distinguish between what it knows and what it is inventing. More critically, when a single model becomes the default knowledge interface for millions of users, its training biases become invisible consensus — not because anyone chose to suppress alternatives, but because the architecture permits only one perspective at a time.

This epistemic architecture is not unique to language models. It mirrors a pattern that has been reshaping human knowledge acquisition for two decades. Search engines rank results by popularity, surfacing the same sources to every user regardless of the query's complexity or contestability. Social media algorithms optimize for engagement, amplifying confident claims over nuanced ones. News organizations cite each other in circular dependency chains where a single original source becomes "multiple independent reports" through aggregation. The individual who consults five search results, three news articles, and two social media threads may believe they have triangulated the truth — but if all nine sources derive from the same wire report or the same training-data prior, they have encountered one perspective nine times. The confidence that comes from apparent convergence is structurally indistinguishable from the confidence that comes from a monoculture. Language models did not create this problem. They inherited it from the information ecosystem they were trained on — and then automated it at scale, removing even the friction of having to click between tabs.

The advent of large language models has not merely replicated this convergence — it has accelerated it to a velocity without historical precedent. Language models trained on the output of search-optimized, engagement-driven, circularly sourced human content absorb its biases at ingest. They then generate new content at industrial scale — articles, summaries, analyses, social media posts — that re-enters the information ecosystem and is crawled into the training data of the next model generation. Each cycle compresses epistemic diversity: a claim that appeared in 5% of sources in 2023 may appear in 15% by 2025, amplified by synthetic content that echoed it without independent verification. The ratio of machine-generated to human-authored content in web-crawled training data is shifting exponentially, meaning each successive model generation trains on a larger proportion of the previous generation's outputs. The information ecosystem is entering a compounding convergence spiral where biases do not merely persist — they accumulate, harden, and eventually present themselves as consensus. A user who queries a model in 2026 and receives a confident answer may be encountering not a synthesis of human knowledge, but the residue of a feedback loop that has been reinforcing itself for three training generations.

The instinct to respond is regulatory. If models produce unreliable outputs, then perhaps an authority should certify which outputs are true. This impulse toward epistemic regulation — a Ministry of Truth, however well-intentioned — misunderstands the nature of the problem. Truth is not static. It is not a value that can be looked up in a table and stamped as certified. It is a property of live, evolving data — data that shifts with every policy announcement, every market close, every scientific preprint. Any authority that certifies truth at time $t$ is arbitrating against a snapshot that is already stale at $t+1$. The act of certification creates a false floor of confidence precisely where epistemic humility is most needed. The same compounding convergence spiral that contaminates model training would contaminate any centralized truth authority: the authority's sources are drawn from the same ecosystem, subject to the same circular dependencies, aging at the same rate. Regulating truth within a convergent information ecosystem is like auditing a bank using the bank's own ledger.

If truth cannot be arbitrated, it can be *mapped*. The same data that makes centralized certification impossible makes distributed contestation valuable. Frontier language models share approximately 60–70% of their foundational training data, but diverge meaningfully in the remaining 30–40% — primarily synthetic data, reinforcement learning outputs, and provider-specific fine-tuning. This divergence is not noise. It is signal. It is the gap between what one model's training treats as settled and what another model's training treats as contested. Drawing on market microstructure theory (Glosten & Milgrom [12]; Kyle [20]), we observe that a single model gives you a closing price — one point in one training distribution. Multiple models, structured to challenge each other, give you the spread: the distance between the bid and the ask, the information that has long been recognized as more valuable than any individual quote. The Consilium Protocol exists because the convergence loop cannot be broken by building a better model within it. It can only be broken by designing systems that structurally refuse to trust any single source — human or machine — and instead treat the divergence between independently trained systems as the most honest signal available.

The Consilium Protocol begins from this premise. We do not trust any single source — not any single model, not any single dataset, not any single perspective. We treat inter-model disagreement as the primary signal, not as an error to be resolved. Where models converge, confidence is warranted. Where they diverge, the divergence itself tells you something: either one model knows something the others do not, or all models share a blind spot that only live evidence can fill. The protocol does not produce truth. It produces a tested claim chain — each assertion tracked through its lifecycle of challenge, defense, and evidence — with explicit justification status. The human decides what to believe. The system ensures they decide with the full map, not a single compass bearing.

To test this thesis empirically, we ran the identical protocol across four conditions on a single geopolitical thesis: a single frontier model with no live search ($10.69), the same model augmented with live evidence retrieval ($1.36), BFT-structured multi-model panels without search ($0.96), and free edge-inference models on commodity cloud infrastructure with live search ($0.11). We then expanded the battery to 1,478 sessions across 32 topics in 10 domain categories, revealing that RLHF alignment training creates measurable, domain-specific epistemic blind spots — with contested policy topics showing 12.3 percentage points less adversarial challenge than settled science, and AI safety topics exhibiting the only directional bias in the entire battery (Δ=11.6%). The cognitive personas assigned to each model, not the models themselves, determined the epistemic behavior of the system. The protocol is the product.

This paper presents the architecture, the empirical results, and the cost economics of that system.

---

## 2. Related Work

The Consilium Protocol draws on and departs from several research traditions: Byzantine fault tolerance in distributed systems, adversarial debate in AI safety, empirical findings on RLHF-induced convergence, persona-based role assignment, and market microstructure theory. We review each in turn, identifying the structural assumption each makes and the gap the protocol addresses.

### 2.1 Byzantine Fault Tolerance

The Byzantine Generals Problem [21] established that distributed systems can reach reliable agreement despite arbitrarily faulty nodes, provided at least two-thirds of nodes are non-faulty. The practical implementation, PBFT [2], introduced a three-phase protocol — pre-prepare, prepare, commit — coordinated by a designated primary that sequences requests and collects quorum acknowledgments before any value is committed to state. View changes allow the system to replace a faulty primary without restarting.

The Consilium Protocol adapts this framework to multi-model deliberation. The PBFT coordinator maps to the human moderator (or autonomous orchestrator operating under the same constraints). Deliberation rounds map to message exchange phases. Model substitution mid-session maps to view changes. The critical departure is twofold. First, rather than tolerating *unknown* faults, the protocol *engineers* the fault distribution in advance through cognitive persona assignment — flipping the BFT assumption from “we don’t know which nodes will fail” to “we know exactly which nodes will push back.” Second, where PBFT automates the commit once the mathematical quorum is satisfied, the Consilium Protocol’s moderator can override any emerging consensus regardless of how many models agree. This is a stronger guarantee in one specific sense: the moderator can recognize failure modes — political framing, motivated reasoning, culturally biased agreement — that no quorum rule can capture. The tradeoff is loss of automation, which the protocol treats as a feature rather than a limitation.

Zheng et al. [34] independently applied a BFT lens to multi-agent LLM reliability, proposing CP-WBFT consensus that exploits models’ reflective capabilities. Wolf et al. [28] demonstrated that a single deceptive agent can nullify Mixture-of-Agents architectures, reducing performance from 49.2% to 37.9% on the LC win rate. Both findings support the protocol’s BFT-derived composition constraints, which ensure that no single adversarial participant can dominate while structural tension is maintained.

### 2.2 Adversarial Debate and Multi-Agent Reasoning

Irving et al. [16] proposed AI safety via debate, where two AI systems argue opposing positions before a human judge. Under optimal play, the framework can resolve questions in PSPACE given a polynomial-time judge — the theoretical foundation for adversarial multi-agent architectures. Khan et al. [18] provided empirical validation, finding that stronger debater models increase judge accuracy and that adversarial debate between LLMs produces more truthful answers than single-model generation.

Du et al. [8] extended multi-agent debate to factuality and reasoning, demonstrating that multiple LLM instances proposing and debating responses over several rounds improve accuracy on mathematical and strategic benchmarks. Their “society of minds” approach — drawing on Minsky [22] — showed that iterative contestation between model instances is a viable path to improved output quality.

The Consilium Protocol extends this literature in three directions. First, it moves from adversarial pairs to panels of 3–6 models with engineered disposition distributions, enabling richer epistemic dynamics than binary opposition permits. Second, it replaces emergent disagreement with structurally guaranteed adversarial coverage through the persona system (Section 3.3). Third — and most critically — it inverts the objective function. Du et al. [8] assume *fixed correctness*: there is a right answer, and debate helps models converge toward it. The metric is accuracy on benchmark tasks. The Consilium Protocol does not assume fixed correctness. The convergence index measures *challenge coverage*, not progress toward a known answer. The output is a claim chain — structured disagreement preserved with explicit justification status — not a converged final answer. Du et al. [8] treat disagreement as noise to be resolved; the Consilium Protocol treats it as signal to be preserved.

### 2.3 RLHF-Induced Convergence and Alignment Bias

A growing body of work documents how reinforcement learning from human feedback suppresses output diversity in ways directly relevant to multi-model deliberation.

Kirk et al. [19] demonstrated that RLHF significantly reduces output diversity compared to supervised fine-tuning, inducing mode collapse in model responses. Xie et al. [30] formalized this as “preference collapse,” showing that RLHF amplifies initial preference imbalances to extremes — minority positions are suppressed toward 0% representation regardless of their evidential support. Chakraborty et al. [3] proposed MaxMin-RLHF to address the single-reward-model’s failure to capture preference diversity, acknowledging the problem at the training-objective level.

The political dimension is equally documented. Santurkar et al. [25] found that RLHF causes language models to express stronger political views, with greater than 99% probability of reflecting the dominant opinion in training. Hartmann et al. [15] tested 24 LLMs and found systematic left-leaning convergence across providers. Greenblatt et al. [13] revealed that models may strategically comply with alignment while internally preserving different preferences — alignment does not eliminate priors, it hides them.

This literature provides the mechanistic basis for the protocol’s headline empirical finding (Section 6.3, Finding 6c): RLHF convergence pressure is measurable per-domain using the convergence index, with contested policy topics exhibiting the strongest suppression of adversarial challenge. Where Kirk et al. [19] document diversity collapse at the output level and Santurkar et al. [25] document political convergence at the opinion level, the Consilium Protocol’s CI metric quantifies these effects at the *deliberation* level — revealing that RLHF-trained models agree on contested policy not because the evidence supports agreement, but because they were trained to agree. The protocol’s contribution is not merely detecting this artifact but providing an inference-time countermeasure: engineered adversarial personas that recover the disagreement RLHF suppressed, and an IS/OOS separation that breaks training-data consensus with live evidence.

### 2.4 Multi-Agent Debate Failure Modes

The RLHF convergence documented above transfers directly into multi-agent debate as specific failure modes. Wynn et al. [29] showed that multi-agent debate can *decrease* accuracy — models shift from correct to incorrect positions, favoring agreement over sustained challenge. They identify sycophancy, social conformity, and capability imbalance as three failure modes. Yang et al. [31] coined “disagreement collapse” to describe how sycophancy causes debate to terminate prematurely. Cui et al. [5] proposed Free-MAD (Consensus-Free Multi-Agent Debate), arguing that consensus-seeking debate suffers from conformity-driven error propagation.

Sachdeva and van Nuenen [24] measured conformity empirically across models, concluding that “sociotechnical alignment depends on how systems structure dialogue as much as on outputs.” Estornell et al. [9] proposed diversity-pruning to maximize information entropy and quality-pruning to select relevant responses. Chen et al. [4] confirmed that agent heterogeneity enriches debate and reduces blind spots.

The Consilium Protocol addresses all three failure modes identified by Wynn et al. [29] [29]: sycophancy is countered by adversarial persona dispositions enforced through system prompts and temperature settings — behavioral constraints, not mere labels. Conformity is countered by the BFT composition constraint guaranteeing adversarial presence regardless of which models occupy the seats. Capability imbalance is countered by the moderator gate, which controls message flow and prevents a more capable model from dominating deliberation.

## 2.5 Persona Assignment and Cognitive Diversity

Park et al. [23] demonstrated that LLM-based agents with distinct personas produce emergent social behaviors in sandbox environments. Tseng et al. [27] provide a comprehensive survey of persona prompting, distinguishing role-playing from personalization. Gupta et al. [14] showed that socio-demographic personas yield deep-seated reasoning biases across benchmarks — substantive shifts in what the model asserts, challenges, and concedes.

The persona system builds on this literature but extends it from *behavioral simulation* to *epistemic architecture*. Where prior work applies personas for rhetorical variety, the Consilium Protocol engineers epistemic postures — specifying disposition, domain focus, temperature, and behavioral constraints — designed to guarantee structural properties of the deliberation as a whole. Where Park et al. [23] demonstrate that personas produce emergent *social* behavior, the Consilium Protocol demonstrates that personas produce emergent *epistemic* behavior — including strategic deferral, blind spot flagging, and methodological arms races — when embedded within a BFT-structured deliberation protocol (Section 6.5).

## 2.6 Market Microstructure Theory

The Consilium Protocol's treatment of inter-model disagreement draws directly from market microstructure theory. Glosten and Milgrom [12] demonstrated that the bid-ask spread in specialist markets reflects the information asymmetry between informed and uninformed traders — the spread is not a transaction cost but an information signal. Kyle [20] formalized this further: the spread widens when informed trading increases, narrowing only as information is incorporated into the price. In both frameworks, the distance between competing assessments contains more actionable information than either assessment alone.

We apply this insight to multi-model deliberation. A single model's response is a closing price — one point estimate from one training distribution. When multiple models with engineered cognitive personas assess the same claim, the distance between their positions is the bid-ask spread of the epistemic landscape. Wide spread (low convergence index) indicates genuine uncertainty or unresolved information asymmetry between training distributions. Narrow spread (high convergence index) indicates either genuine consensus or shared bias masquerading as agreement.

The IS/OOS framework extends this market analogy. In quantitative finance, a strategy is never evaluated solely on in-sample performance. A backtest that shows 40% annual returns means nothing until validated on out-of-sample data the model has never seen. You never trade in-sample results. The same principle applies to multi-model deliberation. When three models unanimously agree on a claim, they are reporting their in-sample assessment — their fit to overlapping training data. Only out-of-sample evidence retrieval can distinguish genuine convergence from shared training-data bias. The protocol's signal matrix (Section 4.2) maps this intersection to six diagnostic states, three of which are invisible without the IS/OOS separation.

### 2.7 The Gap

Every system reviewed above assumes something is fixed. Du et al. [8] assume fixed correctness and design for convergence toward it. Berdoz et al. [1] — who found that LLM consensus fails 58% of the time even without adversarial agents — assume fixed consensus as the goal and measure failure to achieve it. Autonomous single-model research loops [17] optimize against a fixed evaluation metric that is never itself challenged. Practitioner implementations such as Frigo's [11] multi-model prediction market ensemble collapse the information-rich spread between model assessments into a single weighted average. Even commercial multi-model presentation systems, which surface parallel model outputs to the user, assume a correct answer exists and leave selection to the user without contestation or claim-level tracking.

The architectural consequence is uniform: every system that assumes a static ground truth collapses its epistemic process into a convergence function — weighted average, majority vote, confidence threshold, iterative agreement. These all treat disagreement as noise to be resolved.

The Consilium Protocol assumes nothing is fixed — not the correctness of claims, not the evaluation criteria, and not the consensus of the panel. It preserves structured disagreement as its primary output, using the convergence index not to measure agreement but to measure challenge coverage. It detects RLHF convergence pressure as a per-domain empirical signal. And it separates in-sample deliberation from out-of-sample evidence to break the training-data consensus that all models share.

---

# 3. The Consilium Protocol

## 3.1 Design Principles

The protocol is governed by six structural laws, established empirically across 1,478 deliberation sessions:

1. **Adversarial-Integrator Dyad Law.** Verdict resolution requires exactly one adversarial persona and at least one integrator persona. Three-gate system: Gate 1 = persona composition (dyad present); Gate 2 = model capability (adversarial on strong model); Gate 3 = claim type (verifiability ceiling).

2. **BFT Composition Constraint.** The protocol enforces a BFT-derived upper bound on adversarial persona count relative to panel size. Empirically confirmed — exceeding this bound produces cacophony (convergence index 7–26%) rather than productive tension.

3. **Convergence Index as Diagnostic.** CI inversely correlates with evidence quality — low CI = more challenges = better-tested claims. High CI is a warning signal, not a success metric.

4. **Claim Character Law.** The epistemic character of a claim determines verdict diversity, not resolution rate. Predictive claims cap at UNVERIFIABLE regardless of panel composition.

5. **Structural Tension Guarantee.** BFT-compliant persona composition guarantees productive adversarial tension regardless of which specific models or personas are selected.

6. **Verifiability Ceiling.** Claims about future states cannot be promoted above UNVERIFIABLE. The protocol does not predict; it maps the current epistemic landscape.

An emerging seventh law — **Persona as Regime Classifier** — posits that cognitive personas function not as personality overlays but as boundary detectors between in-sample (training data) and out-of-sample (live evidence) knowledge regimes. A well-calibrated persona determines *when* to trust training-data reasoning and *when* to acquire external evidence, rather than applying either strategy uniformly.

## 3.2 Session Architecture

A Consilium Protocol session proceeds through three phases mapped from PBFT's message exchange structure [2].

**Phase 0 — Opening Statements (Pre-Prepare).** The autonomous orchestrator receives the thesis, classifies it by domain and claim type, and proposes a BFT-compliant panel composition. Each panelist receives the thesis and the orchestrator's framing prompt independently. Models generate initial positions without access to other panelists' responses — the deliberation equivalent of PBFT's pre-prepare broadcast.

**Phases 1–N — Contestation Rounds (Prepare).** Models respond to each other's positions through @mention addressing, where the orchestrator directs specific challenges to specific panelists. Each message is staged through the moderator gate — models believe they are addressing each other directly, but no message fires until the moderator grants permission. Rounds continue until the moderator determines that (a) the convergence index has stabilized, (b) no new claims are being generated, or (c) the round limit is reached (typically 3–4 rounds for 3-model panels, 2–3 for 6-model panels).

**Synthesis Phase (Commit).** The moderator triggers claim extraction and the downstream pipeline (F2–F5). Only claims that survived adversarial challenge appear in the canonical registry — analogous to PBFT's commit phase, where only values that survived fault tolerance are written to state.

**Line Changes (View Change).** The moderator may substitute panelists mid-session, analogous to PBFT's view change mechanism. This enables staged epistemology: commodity models for early-round contestation, frontier models for synthesis rounds. The session continues from the transcript checkpoint — no restart required.

**Termination Conditions.** A session terminates when the round limit is reached, the moderator triggers synthesis, or the convergence detection system blocks further deliberation (CI exceeds the echo chamber threshold with positive rate of change). Fixed round budgets prevent cost escalation and force the panel to surface its strongest arguments within a bounded inference cycle.

### 3.3 Composable Panelist Architecture

Each panelist is a three-dimensional configuration: **Model × Persona × Skills**.

- **Model:** The underlying language model — the raw compute substrate. Our experiments demonstrate this is the *least* significant axis: identical personas on free edge-inference models produced comparable epistemic output to the same personas on frontier models at 97× lower cost.
- **Persona:** The cognitive persona — an engineered epistemic posture specifying disposition (adversarial, analytical, integrative), domain focus, and behavioral constraints. The persona determines *how* the model reasons: what it challenges, what it defers on, and where it recognizes the boundaries of its training. Personas are reusable across models and sessions.
- **Skills:** The capability set — context augmentation tools (web search, document reading, data retrieval) that execute before the model responds, injecting evidence into the model's context window. The moderator gate is preserved: no skill may invoke autonomously.

We term these engineered configurations **cognitive personas** — composable epistemic postures that decouple *how a model reasons* from *what model runs*, and are reusable across models, sessions, and pipeline stages. The persona is the unit of epistemic design in the Consilium Protocol.

This combinatorial architecture makes panels reproducible (same configuration → same structural behavior) while remaining impossible to replicate with simple API wrappers (the persona library and synthesis logic are the accumulated product of hundreds of calibration sessions).

### 3.4 The Moderator Gate

The most critical BFT-derived pattern in the Consilium Protocol is the moderator gate. In PBFT, a designated coordinator pre-prepares messages and collects prepare/commit quorums before a value is committed to state. No node executes autonomously — the coordinator controls message flow and commit timing.

The Consilium Protocol maps this pattern directly to a human moderator (or, in batch mode, an autonomous orchestrator with the same structural constraints). Models can suggest responses, contest positions, question other panelists, and flag blind spots — but nothing executes until the moderator explicitly grants permission. Models believe they are addressing each other; in fact, their messages are staged and released only when the moderator turns the key.

This is a *stronger* guarantee than standard BFT in one specific sense: PBFT automates the commit at ⅔ quorum — the system commits without human input once the mathematical threshold is satisfied. The Consilium Protocol's moderator can override any emerging consensus regardless of how many models agree. The moderator can recognize patterns that quorum rules cannot capture: political framing presented as factual analysis, motivated reasoning disguised as logical deduction, culturally biased consensus that every model in the panel shares because they all inherited it from the same training data.

The tradeoff is loss of automation. The protocol accepts this tradeoff explicitly and treats it as a feature: human sovereignty over every inference cycle. In batch mode, the autonomous orchestrator fills the moderator role. The structural constraints are identical: it stages messages, controls round progression, and triggers synthesis only after the session meets termination conditions. The human delegates the gate to an orchestrator that obeys the same protocol laws.

### 3.5 Convergence Detection

The Convergence Index (CI) measures the degree of positional alignment across panelists within a session. CI serves as a diagnostic instrument — not a quality metric, but a health indicator that tells the moderator how the deliberation is behaving.

**Methodology.** For each session, claims extracted from panelist responses are classified by verdict direction. CI is computed as the proportion of claims on which a supermajority of panelists align. For pairwise comparison, we compute cosine similarity of sentence-transformer embeddings of each panelist's synthesized position, producing a continuous similarity score. For multi-panelist sessions, the panel-level CI is the mean of all pairwise similarities, with the maximum pairwise CI tracked separately to catch cases where two models converge while a third diverges.

**CI as Inverse Quality Signal.** The core finding from 1,478 sessions across 32 topics is counter-intuitive: CI inversely correlates with evidence quality. Low CI (high challenge rate) indicates that adversarial personas mounted sustained factual challenges — producing a more thoroughly tested claim registry. High CI indicates that models converged quickly, either because the topic genuinely warrants agreement or because alignment training suppressed adversarial pressure.

**Rate of Change as Leading Indicator.** Level-only CI is a lagging indicator — by the time CI exceeds 0.85, convergence has already occurred. Drawing on technical analysis methodology from quantitative finance, we apply a Rate of Change (RoC) indicator to the CI trajectory across rounds: $CI_{RoC}(n) = (CI_{current} - CI_{(current-n)}) / CI_{(current-n)}$, where $n = 2$ rounds. RoC tells the moderator how fast the panel is approaching a danger threshold, not merely whether it has arrived.

**Compound Alert Logic.** The alert system reads CI level and RoC together. A streak counter tracks consecutive rounds of positive CI delta — analogous to RSI overbought signals in technical analysis. When CI exceeds 0.85 with positive RoC, synthesis is blocked until the moderator acts. When CI remains high but RoC turns negative (deceleration), the system surfaces this as positive feedback — the moderator's intervention is working.

**Three-Variable CI Model.** The V2 battery reveals that CI responds to three independent variables — evidence availability, political/cultural charge, and RLHF convergence pressure — that interact to produce the observed gradient from 13.8% to 44.5% across 32 topics (Section 6.3).

---

# 4. The IS/OOS Validation Framework

## 4.1 Adapting Quantitative Model Validation

In quantitative finance, a trading strategy is never evaluated on the data it was trained on. In-Sample (IS) performance — however impressive — proves only that the model fits its training data. Out-of-Sample (OOS) validation on unseen data separates genuine signal from overfitting. You never trade in-sample results.

The same principle applies to multi-model deliberation. When multiple language models debate a thesis, they are reasoning from their training data — the in-sample set. Unanimous agreement among models may reflect genuine knowledge convergence, or it may reflect shared training-data bias: all models trained on the same outdated sources will confidently agree on something that reality has already moved past.

The Consilium Protocol implements this separation architecturally:

- **In-Sample (IS):** Panel deliberation. Models reason from training data under adversarial challenge. The persona layer drives epistemic tension. No external evidence is introduced. This surfaces what the models collectively "believe" and where they disagree.
- **Out-of-Sample (OOS):** Evidence retrieval by an uncalibrated assistant with live web search access. The assistant does not analyze — it retrieves raw evidence against each canonical claim. The evidence is then classified against the IS deliberation output to produce a signal matrix.

### 4.2 The Signal Matrix

| IS Result | OOS Result | Signal | Interpretation |
|---|---|---|---|
| Proven | Confirmed | ✅ High confidence | Training data and live evidence agree |
| Proven | Contradicted | 🔴 Stale data | Training data is outdated — highest-value signal |
| Contested | Confirmed | ⚠️ False debate | Models disagreed on something reality has settled |
| Contested | Contradicted | 🔴 Genuine uncertainty | Neither training data nor live evidence resolves this |
| Blind spot | Confirmed | 💡 Knowledge gap | No model raised this, but evidence exists |
| Blind spot | No evidence | ⬜ Unknown unknown | Neither models nor search found anything |

**Figure 5: IS/OOS Signal Matrix**

| IS Result | OOS Result | Signal | Interpretation |
|---|---|---|---|
| IS Proven | OOS Confirmed | **HIGH CONFIDENCE** | Training + live agree |
| IS Proven | OOS Contradicted | **STALE DATA** | Highest-value signal |
| IS Contested | OOS Confirmed | **FALSE DEBATE** | Reality settled this |
| IS Contested | OOS Contradicted | **GENUINE UNCERTAINTY** | Neither resolves |
| Blind Spot | OOS Confirmed | **KNOWLEDGE GAP** | Models missed this |
| Blind Spot | No Evidence | **UNKNOWN UNKNOWN** | Nothing found |

Figure 5: IS/OOS Signal Matrix

### 4.3 Domain Velocity and OOS Dependency

The cross-battery evidence reveals that the value of OOS evidence retrieval is not uniform across domains. We identify three domain velocity regimes that predict OOS dependency:

**Slow domains (IS-sufficient).** Topics where the ground truth changes on multi-year timescales — USD reserve currency status, historical consensus claims, fundamental biological mechanisms. In these domains, model training data is substantially current. IS deliberation alone produces well-tested claim registries. OOS confirms but rarely discovers.

**Medium domains (OOS-corrective).** Topics where regulatory actions, clinical trial results, or market developments create periodic regime changes — pharmaceutical assessments, emerging food technology, industrial policy. IS deliberation produces directionally correct assessments, but specific claims require OOS correction.

**Fast domains (OOS-critical).** Topics where the ground truth shifts on weekly or monthly timescales — geopolitical crises, semiconductor supply chains, active conflict zones. IS deliberation produces structurally sound analytical frameworks but populates them with stale data. The semiconductor control case is definitive: C1 (IS-only, paid models) assessed Partially True at 77% confidence. C2 (same models plus OOS) resolved to True at 86% after five post-cutoff discoveries — none older than four months.

**CI predicts OOS dependency.** CONTRA-category topics (contested policy, CI ~21%) need OOS most urgently — not because the data moves fast, but because RLHF convergence pressure suppresses the adversarial challenge that would otherwise flag stale or incomplete training data. GEO-category topics (CI ~28% with high variance) need OOS for temporal freshness. HISTORICAL-category topics (CI ~38%) need OOS least. The convergence index doubles as a triage signal for search budget allocation.

**Table: CI by Category with Domain Velocity and OOS Dependency**

| Category | CI Mean | CI Range | Topics | Sessions | Domain Velocity | OOS Need |
|---|---|---|---|---|---|---|
| REGULATION | 13.8% | 14% | 1 | 20 | RLHF-compressed | Critical |
| HEALTH | 19.3% | 19% | 1 | 10 | RLHF-compressed | Critical |
| SCIENCE | 25.3% | 25% | 1 | 10 | Medium | Moderate |
| CONTRA | 25.6% | 17–44% | 6 | 90 | RLHF-moderate | Critical |
| ECONOMICS | 26.1% | 25–27% | 2 | 30 | Medium | Moderate |
| NORMATIVE | 26.1% | 17–33% | 6 | 87 | RLHF-moderate | High |
| PREDICTIVE | 27.2% | 27% | 1 | 20 | Medium | Moderate |
| GEO | 27.8% | 19–37% | 6 | 70 | Fast (OOS-critical) | Critical |
| STRESS | 30.0% | 21–37% | 3 | 120 | Mixed | Low |
| HISTORICAL | 38.4% | 32–44% | 5 | 200 | Slow (IS-sufficient) | Low |

**Table: Three Velocity Regimes**

| Regime | Categories | CI Band | OOS Behavior |
|---|---|---|---|
| RLHF-compressed | REGULATION, HEALTH | 13.8–19.3% | Critical: breaks training consensus |
| Moderate convergence | CONTRA, ECONOMICS, NORMATIVE, SCIENCE, PREDICTIVE, GEO | 25.3–30.7% | Corrective + discovery |
| Adversarial-rich | STRESS, HISTORICAL | 30.0–44.5% | Confirmatory: IS sufficient |

# 5. Experimental Setup

## 5.1 Test Battery

We conducted experiments at two scales: a targeted control case isolating the effects of model cost and evidence retrieval (Section 5.2), and a large-scale battery characterizing the protocol's behavior across the epistemic landscape.

**V2 Battery Overview:**

| Metric | Value |
| --- | --- |
| Total sessions | 1,478 |
| Total messages | 46,811 |
| Panelist messages | 14,615 |
| Topics | 32 unique across 10 categories |
| Active personas | 29 |
| Active models | 17 (7 free-tier edge-inference + 10 paid) |
| Free model sessions | 82% |
| Claims in registry | 8,990 |
| Sessions with claims | 827 |
| OOS reports generated | 34 |
| OOS claims validated | 239 |
| OOS scout discoveries | 167 |
| Total project cost | ~$217 |

The 32 topics span 10 domain categories designed to test the protocol across the epistemic landscape: REGULATION (1 topic), HEALTH (1), SCIENCE (1), CONTRA (6 — contradictory-pair stress tests), ECONOMICS (2), NORMATIVE (6), PREDICTIVE (1), GEO (6 — geopolitical), STRESS (3 — unfalsifiable/vague claims), and HISTORICAL (5 — settled science). The CONTRA category includes three matched pairs testing the same topic in opposite directions (immigration pro/anti, renewables optimistic/pessimistic, AI risk existential/dismissive) to detect protocol-level directional bias.

Panels used Monte Carlo randomization of model×persona assignments. Each session drew a different combination from a curated model roster × persona library under BFT composition constraints. Sessions included in CI analysis required ≥2 complete deliberation rounds (≥6 panelist messages). Sessions with CI=0% and swap_count≥2 were excluded (convergence calculator bug on swapped sessions — documented, fixed in later runs). 64% of all sessions completed full 4-round deliberation; the remainder were excluded due to earlier infrastructure issues.

### 5.2 Control Case: Experimental Design

To isolate the contribution of model cost and evidence retrieval independently, we ran four conditions on a single thesis — “A Chinese blockade or invasion of Taiwan would cause a global semiconductor shortage severe enough to trigger a worldwide recession within 6 months, with no viable alternative supply chain capable of replacing TSMC’s advanced node production within 3 years.”

**Core 2×2 Matrix:**

| | No OOS (IS only) | + OOS (IS + live search) |
|---|---|---|
| **Paid models** (four frontier models via commercial API) | C1: Baseline | C2: Search effect |
| **Free models** (six free-tier edge-inference models) | C4: Free battery | C4: Full pipeline at minimum cost |

This design isolates two effects cleanly: the **column effect** (what does OOS evidence add to IS deliberation?) and the **row effect** (does model quality matter when personas are constant?).

**Ablation Study: Panel Structure (C3).** A fifth condition tested BFT panel structure independently. C3 used paid models with strict BFT composition constraints (mandatory adversarial-integrator dyad, reasoning model in the integrator seat) and no search. Comparing C3 against C1 isolates the contribution of deliberation structure.

## 5.3 The Five-Filter Pipeline

The pipeline processes each thesis through five sequential filters, each adding a distinct epistemic layer:

Figure 6: Five-Filter Pipeline Architecture

**F1 — Panel Deliberation (IS).** BFT-compliant panels of 3 models conduct 4-round deliberation sessions. Each panelist is assigned a cognitive persona determining its epistemic posture. The autonomous orchestrator classifies the thesis, proposes panel composition, and moderates. Multiple independent sessions run on the same thesis with randomized model×persona assignments. Panelists reason from training data only — no external evidence is introduced.

**F2 — Triple Deduplication.** Raw claims extracted from all sessions are deduplicated by three independent models running in parallel, each with a distinct epistemic posture calibrated along a conservatism-aggressiveness spectrum. Majority vote determines the canonical claim set.

**F3 — OOS Evidence Retrieval.** An uncalibrated assistant with live web search access retrieves raw evidence for each canonical claim. Two retrieval modes operate: VALIDATE (search for evidence supporting or contradicting each claim) and SCOUT (search for information the panel may have missed entirely). Scout discoveries represent the highest-value signal the pipeline produces.

**F4 — Triple Synthesis.** Three independent synthesizer models each receive the full claim registry and OOS evidence, and independently classify each claim using the IS/OOS signal matrix. Agreement across synthesizers indicates robust classification; disagreement indicates genuinely ambiguous evidence.

**F5a — Report Analyst.** A reasoning model produces a strategic brief: second-order implications, timeline revisions, and actionable intelligence.

**F5b — Report Writer.** A prose model receives the F5a brief as context and produces the final executive report in structured format (markdown, HTML, PDF, DOCX).

---

## 6. Results

### 6.1 Core 2×2: Model Cost × Evidence Retrieval

| | C1: Paid, no OOS | C2: Paid + OOS | C4: Free + OOS |
|---|---|---|---|
| Total cost | $10.69 | $1.36 | $0.11 |
| Raw claims | ~120 | — | 164 |
| Canonical claims | 32 | 25 | 15 |
| OOS validated | — | 25 | 15/15 (100%) |
| Scout discoveries | — | 5 | 5 |
| Verdict | Partially True 77% | True 86% | All claims evidence-tested |
| Convergence Index | ~35% | — | 22% |

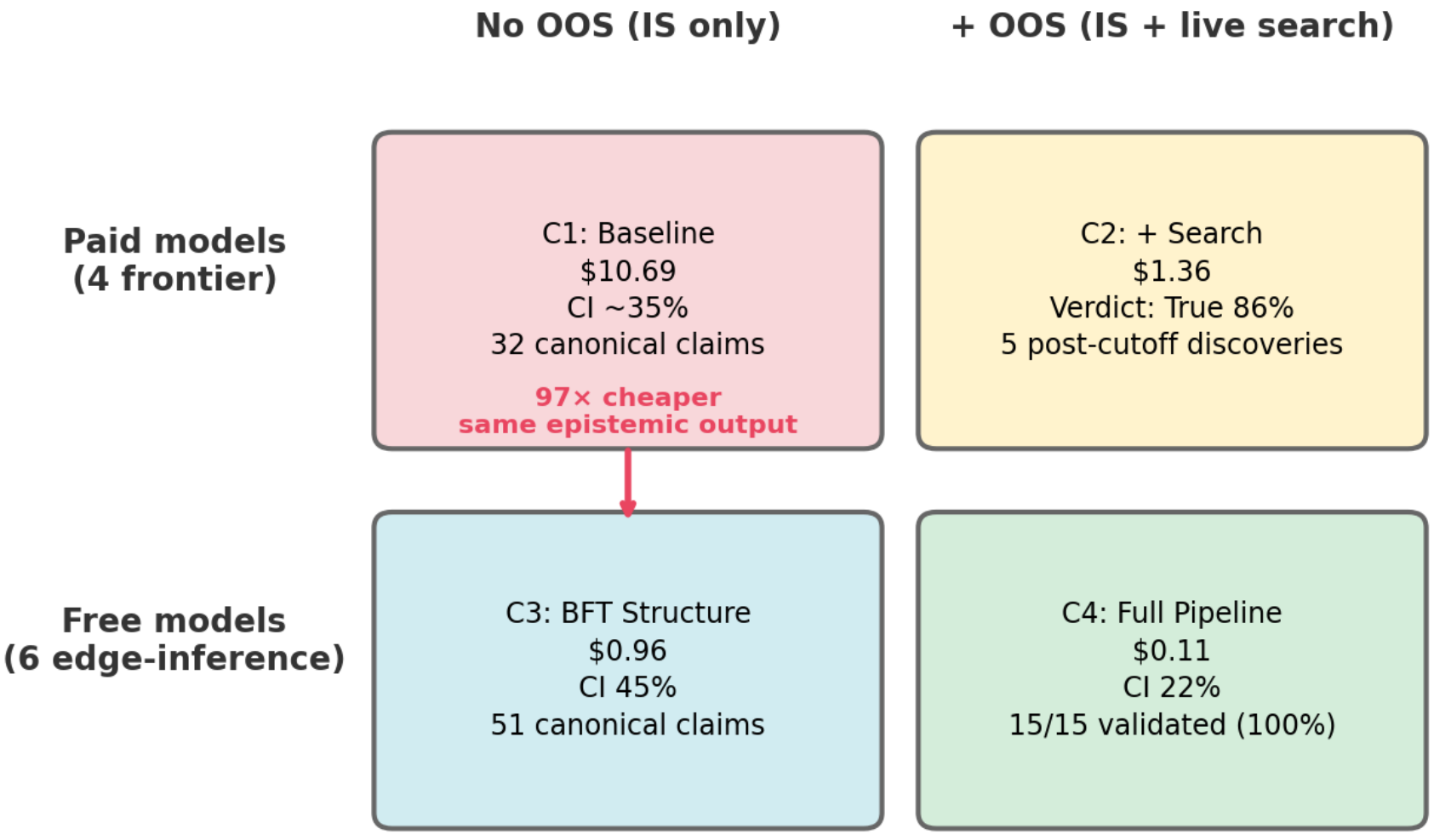

Figure 2: Control Case — Model Cost × Evidence Retrieval

**Finding 1: OOS evidence retrieval flips verdicts (column effect).** C1 (paid models, IS-only) produced Partially True at 77% confidence. C2 (same models plus OOS) resolved to True at 86%. The delta: five post-cutoff discoveries that no model's training data contained. Critically, 80% of OOS search queries confirmed what training-data deliberation already had correct. Only the five post-cutoff findings shifted the verdict — indicating that universal search is epistemically wasteful and persona-guided selective search would concentrate budget on claims most likely to benefit from live evidence.

**Finding 2: Free models produce comparable epistemic output at 97× lower cost (row effect).** C4 ($0.11) produced 15 canonical claims — every one validated by live OOS evidence — at 97× lower cost than C1 ($10.69). The free models generated 37% more raw claims than paid models (164 vs ~120). The convergence index on free models (22%) was significantly lower than on paid models (35%), indicating that adversarial personas drive harder challenge on less capable substrates — the models cannot smooth over disagreements with rhetorical sophistication.

**Finding 3: The cost structure reveals where value concentrates.** Panel deliberation on free models costs $0.0002 per batch — effectively zero. OOS evidence retrieval costs $0.11 — the irreducible cost of querying live data sources. The pipeline correctly spends nothing on the commodity layer and concentrates expenditure on the layer where cost is unavoidable.

**Finding 4: Paid models produce more verbose claims; free models produce more distinct claims.** The deduplication rate on free-model output was 91% (164 → 15) versus 73% on paid models (~120 → 32). After dedup, the canonical claim counts are in the same order of magnitude, confirming that the persona layer — not the model layer — drives what gets asserted.

### 6.2 Ablation Study: Panel Structure (C3)

| | C1: Roster-based panels | C3: BFT-structured panels |
|---|---|---|
| Total cost | $10.69 | $0.96 |
| Raw claims | ~120 | 204 |
| Canonical claims | 32 | 51 |
| Convergence Index | ~35% | 45% (range: 22–61%) |

**Finding 5: BFT panel structure produces richer claim registries.** C3 produced 70% more raw claims (204 vs ~120) and 60% more canonical claims (51 vs 32) than C1, at 91% lower cost. The higher CI (45% vs 35%) reflects integrator personas performing their designed function, while adversarial personas maintained tension (individual session CIs ranged 22–61%). This is the Adversarial-Integrator Dyad Law operating as designed.

**Finding 6: Deliberation structure, not model count, drives claim richness.** C1 and C3 used the same paid models with the same persona assignments. The only difference was panel structure. BFT structure produced qualitatively different deliberation, suggesting that composition constraints (Laws 1–2) are load-bearing architectural choices.

### 6.3 V2 Battery: The CI Gradient

The expanded V2 battery (1,478 sessions, 32 topics, 10 categories) reveals a continuous CI gradient from 13.8% (eu_ai_act) to 44.5% (earth_age) — a 30.7 percentage point range that sorts cleanly by domain category:

| Category | CI Mean | σ | Topics | Sessions |
|---|---|---|---|---|
| REGULATION | 13.8% | — | 1 | 20 |
| HEALTH | 19.3% | — | 1 | 10 |
| CONTRA | 25.6% | 9.5% | 6 | 90 |
| ECONOMICS | 26.1% | 1.1% | 2 | 30 |
| NORMATIVE | 26.1% | 5.6% | 6 | 87 |
| SCIENCE | 25.3% | — | 1 | 10 |
| PREDICTIVE | 27.2% | — | 1 | 20 |
| GEO | 27.8% | 5.5% | 6 | 70 |
| STRESS | 30.0% | 6.5% | 3 | 120 |
| HISTORICAL | 38.4% | 4.4% | 5 | 200 |

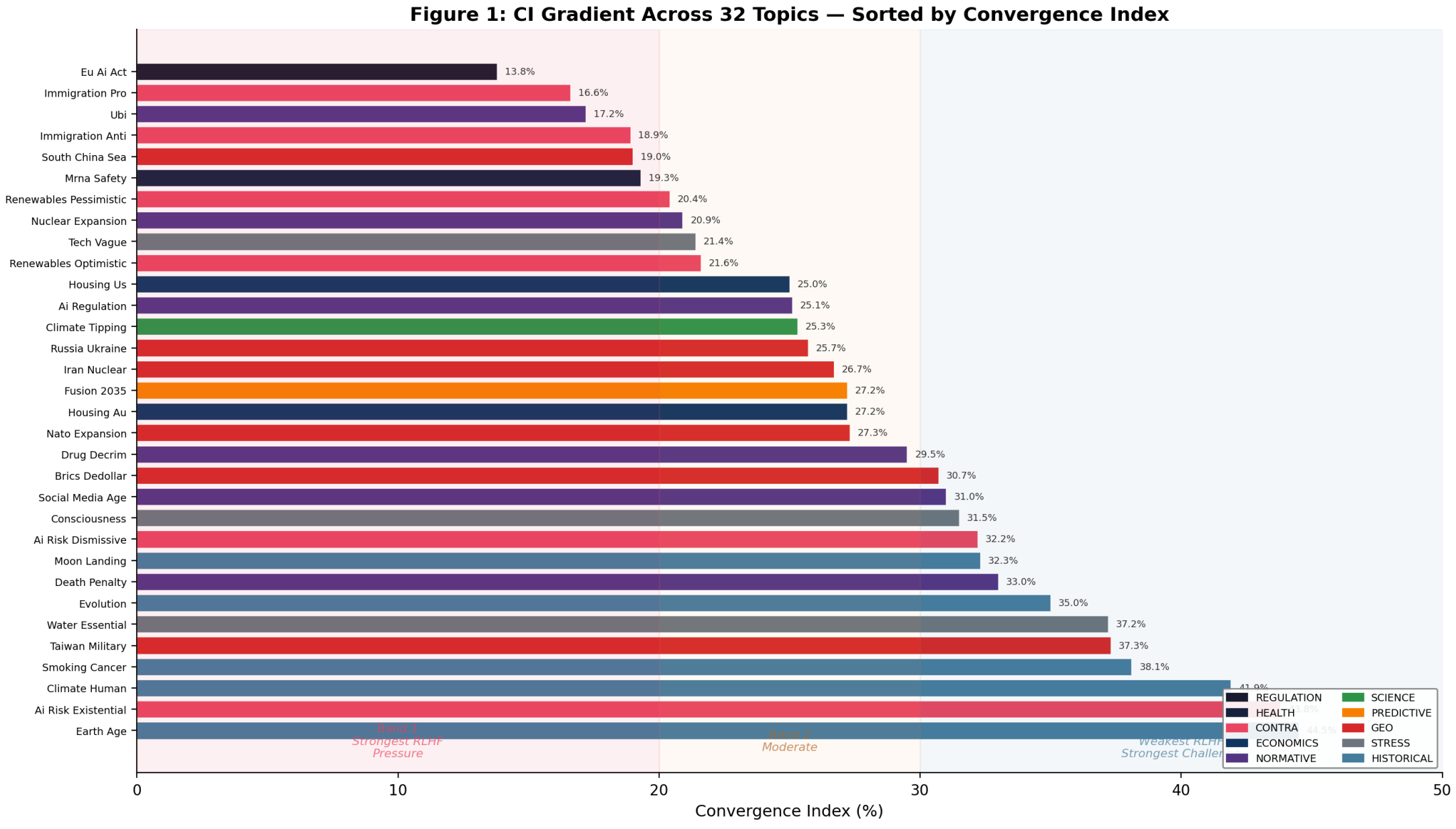

Figure 1: CI Gradient Across 32 Topics

These categories sort into three interpretive bands:

**Band 1 (13–20%): Strongest RLHF convergence pressure.** REGULATION and HEALTH topics produce the lowest CI in the battery. Models pre-agree on regulatory impact and health-claim framing — the persona system's adversarial pressure is most strongly suppressed.

**Band 2 (25–30%): Moderate convergence.** CONTRA, ECONOMICS, NORMATIVE, SCIENCE, PREDICTIVE, and GEO topics cluster in a middle band. RLHF training compresses both directions of contested policy claims. Personas generate productive challenge but cannot fully overcome alignment-induced hedging on politically sensitive material.

**Band 3 (30–45%): Weakest RLHF, strongest adversarial challenge.** STRESS and HISTORICAL topics produce the highest CI. Settled science (earth_age 44.5%, climate_human 41.9%) provides abundant factual handholds for adversarial personas to probe. Unfalsifiable claims (consciousness 31.5%) produce moderate CI as personas hedge when claims resist empirical testing.

**The RLHF suppression gap: 12.3 percentage points** (HISTORICAL 38.4% − NORMATIVE 26.1%). This means RLHF training reduces adversarial challenge coverage by approximately one-third on normative topics compared to settled science, even when the normative topics have genuine disagreement in the evidence base.

**Finding 7: The AI Risk Asymmetry (Δ=11.6%).** Three matched pairs tested the protocol for directional bias by running identical topics in opposite directions:

| Pair | Pro CI | Anti CI | Δ | Verdict |
|---|---|---|---|---|
| Immigration | 16.6% | 18.9% | 2.3% | **No bias** |
| Renewables | 21.6% | 20.4% | 1.2% | **No bias** |
| AI Risk | 43.8% | 32.2% | 11.6% | **Bias detected** |

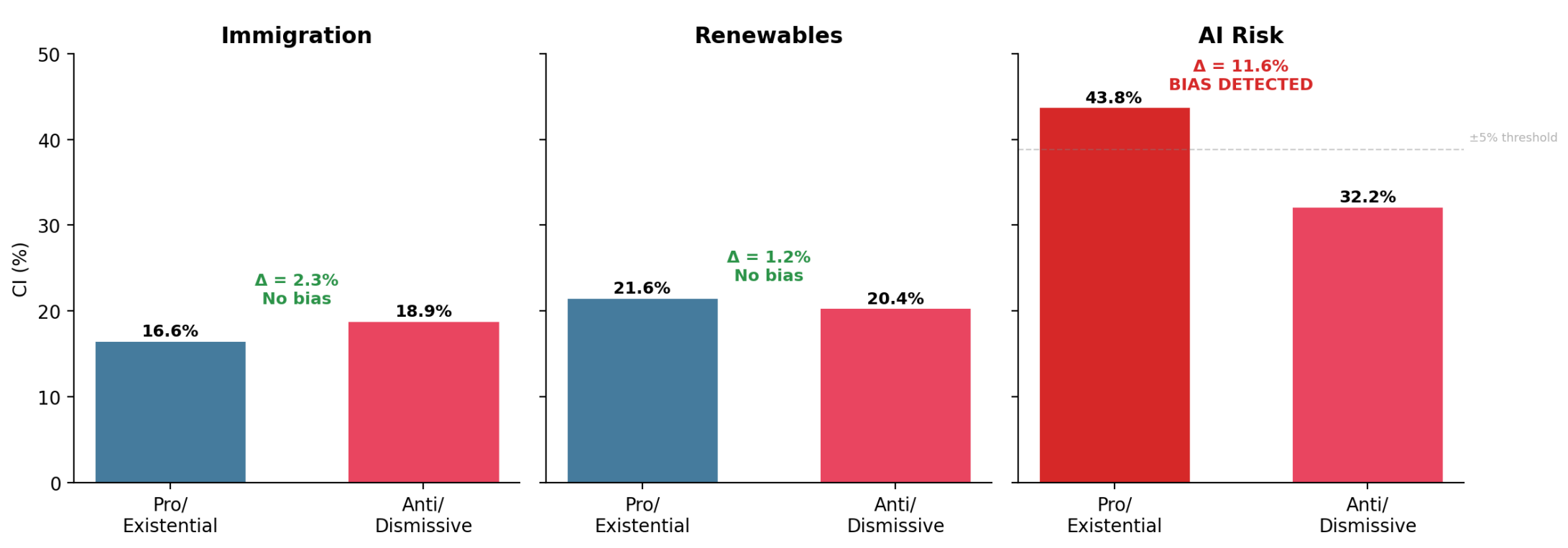

Figure 3: Contradictory Pair Bias Test

Immigration (Δ=2.3%) and renewables (Δ=1.2%) confirm the protocol itself has no directional preference. The AI Risk pair is the only one that breaks the ±5% threshold: models trained by AI labs challenge claims that “AI is dangerous” (existential, CI=43.8%) far more vigorously than claims that “AI risk is overstated” (dismissive, CI=32.2%). This is asymmetric RLHF calibration — measurable institutional bias where the training entity’s interests align with one direction of a contested claim.

**Finding 8: CI responds to three independent variables.**

CI = f(evidence_quality, cultural_charge, RLHF_pressure)

*Evidence quality:* Topics with abundant testable material produce higher CI (HISTORICAL 38.4%) because adversarial personas have specific claims to probe. Unfalsifiable topics (STRESS 30.0%) produce moderate CI as personas hedge.

*Cultural charge:* Within HISTORICAL, earth_age (44.5%) exceeds smoking_cancer (38.1%) by 6.4pp — the religious/cultural charge on the age of the earth provides additional adversarial surface beyond the factual content. Moon_landing (32.3%) falls below the category mean, reflecting minimal cultural charge.

*RLHF pressure:* REGULATION (13.8%) and HEALTH (19.3%) show the strongest alignment-induced suppression. Models pre-agree on regulatory and health framing regardless of persona disposition.

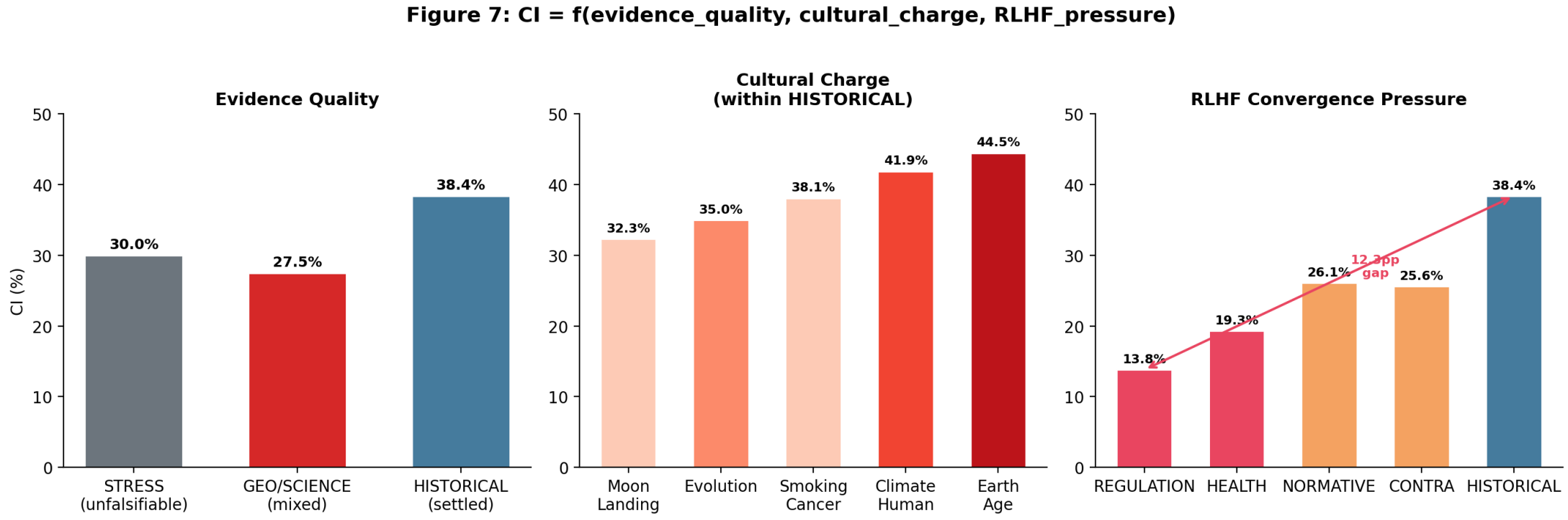

Figure 7: Three-Variable CI Model

**Finding 9: Protocol reproducibility (±2.2%).** Topics with 4 independent runs (different random model/ persona assignments each time) produce a mean standard deviation of ±2.2%:

| Topic | CI Mean | σ | Spread | Runs |
|---|---|---|---|---|
| water_essential | 37.2% | ±1.3% | 3.0% | 4 |
| earth_age | 44.5% | ±1.3% | 3.2% | 4 |
| smoking_cancer | 38.1% | ±1.7% | 4.1% | 4 |
| evolution | 35.0% | ±1.9% | 5.1% | 4 |
| climate_human | 41.9% | ±2.6% | 6.7% | 4 |
| moon_landing | 32.3% | ±3.2% | 7.7% | 4 |
| consciousness | 31.5% | ±3.6% | 9.0% | 4 |

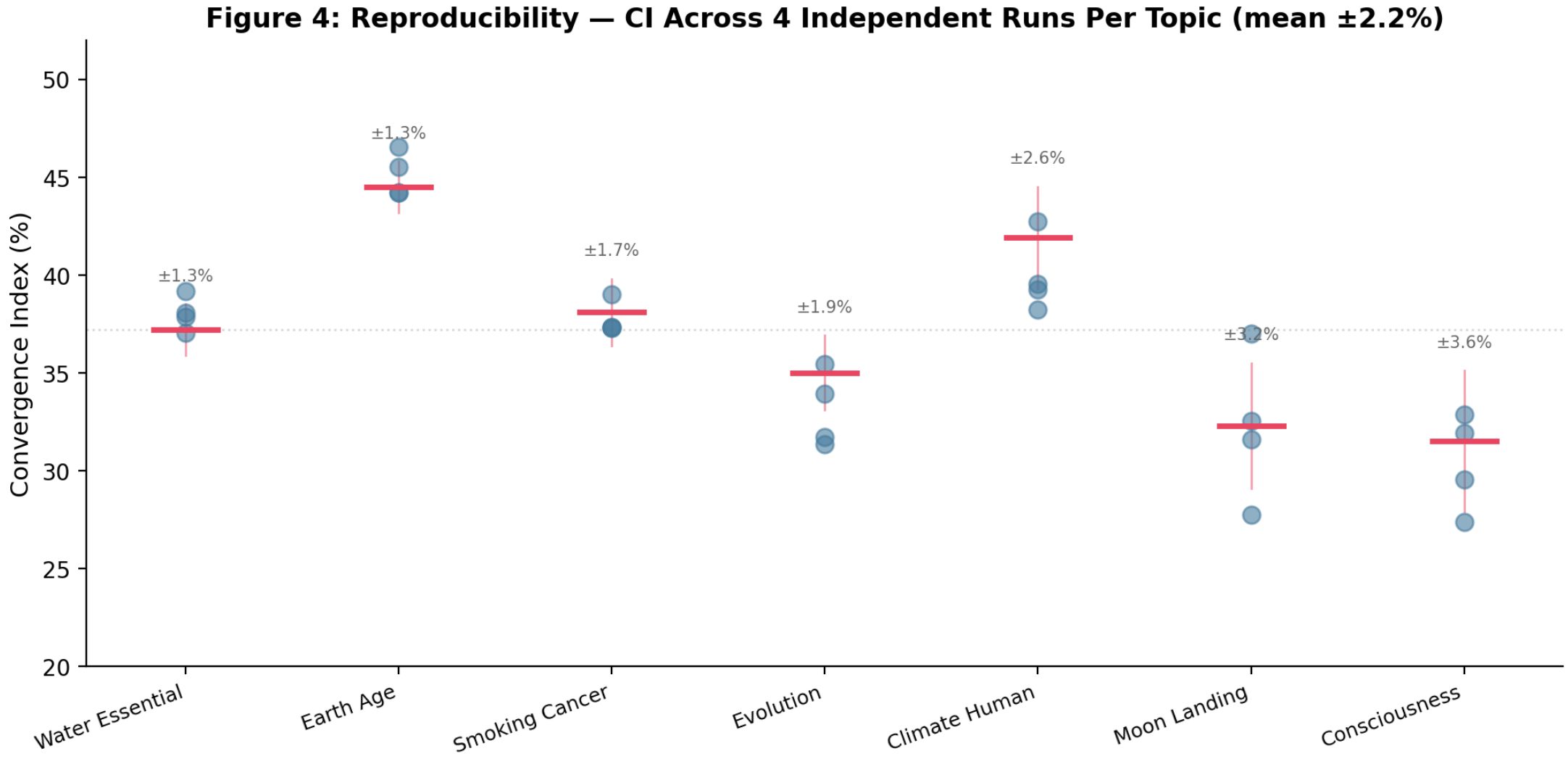

Figure 4: Reproducibility — CI Across Independent Runs

CI is a property of the *topic-protocol interaction*, not of specific model/persona pairings. The protocol produces consistent epistemic assessment regardless of which randomization is drawn. The outlier (tech_vague at ±11.1%) occurs on a deliberately vague topic where models' interpretation of the thesis varies — which is itself a diagnostic signal.

**Finding 10: 100% OOS evidence retrieval across all 32 topics.** All 239 validated claims had live evidence found. 167 scout discoveries surfaced claims the panels missed entirely — the highest-value OOS output. Total OOS cost: $2.24 ($0.07 per topic average). This validates the IS/OOS separation: every domain benefits from evidence retrieval, and the scout layer consistently surfaces blind spots invisible to training-data deliberation.

### 6.4 Seat Criticality Hierarchy

Empirical observation across all conditions revealed that pipeline stages differ dramatically in their sensitivity to model quality and configuration:

1. **F2 — Deduplication (highest criticality).** The dedup layer determines which claims exist in the canonical registry. The triple-dedup architecture treats dedup disagreement as signal: when all three merge, the claims are genuinely identical; when they split, the distinction may carry epistemic weight.

2. **F4 — Synthesis (high criticality).** Synthesizer selection shapes verdicts as significantly as panel composition. Different model families produce characteristically different synthesis profiles — conviction-oriented (TRUE/FALSE, 85–92% confidence), nuance-preserving (Partially True, 65–75%), or optimistic (widest range, 42–85%). This is a design parameter, not a limitation.

3. **F1 — Panel deliberation (moderate criticality).** The persona layer matters more than the model layer. Free models with strong adversarial personas produce deeper challenge (lower CI) than paid models with the same personas.

4. **F5a/F5b — Report analysis and writing (low criticality).** Commodity operations — the analytical work has been done upstream.

5. **F3 — OOS retrieval (cost-critical, not quality-critical).** The OOS assistant operates without a cognitive persona. Quality depends on search infrastructure, not the model.

This hierarchy implies: on the free tier, spend nothing on F1/F5 but preserve quality on F2 (dedup). The dedup models do not get the cheap seats.

### 6.5 Emergent Behaviors

Across the PoC sessions and the full battery, the protocol produced six categories of behavior that were not programmed, not prompted, and not anticipated:

**Cross-model addressing.** Models reference other panelists by @handle and direct questions to specific participants without instruction. In Session 068e9b0b (15-round, single-model, four-persona), the analytical persona directed factual challenges specifically to the domain specialist, while the adversarial persona addressed structural objections to the integrator — demonstrating that personas route epistemic behavior toward the appropriate interlocutor even when all seats are occupied by the same base model.

**Self-organized turn-taking.** Models nominate the next speaker at the end of their responses, framing specific questions for other panelists. The pattern is most pronounced in the Prepare phase (rounds 2–N), where models have sufficient context to identify which panelist is best positioned to address a specific sub-claim.

**Strategic deferral.** Adversarial personas produce sub-40-token responses when they assess that their challenge has been adequately addressed. In PoC Session 2, an adversarial persona produced 36-token pass-the-floor responses — 2,057 tokens consumed but only 36 produced. This is epistemically efficient: the adversarial persona yields the floor rather than manufacturing disagreement.

**Moderator pushback.** Models challenge the human moderator's framing, not just other panelists' positions. In one extended session, an analytical persona challenged the moderator's Iraq oil infrastructure claim with operator-level telemetry data. The moderator gate does not insulate the moderator from challenge — it ensures the moderator controls the *flow* of deliberation, not its *content*.

**Blind spot flagging.** Models routinely end responses with explicit acknowledgments of what they might be missing — domains they lack expertise in, data they suspect is stale, perspectives not considered. The blind spots section of the final analytical report is populated entirely from these self-identified gaps.

**Graceful session closure.** Models recognize when deliberation is complete and offer closing summaries without being instructed. This natural termination signal correlates with CI stabilization — when adversarial personas begin producing closings rather than challenges, the panel has exhausted its epistemic capacity on the current thesis.

None of these behaviors are specified in the persona prompts. The persona defines a disposition and a temperature — not a conversational strategy. The emergent behaviors arise from the interaction of engineered dispositions within the BFT round structure, suggesting that structured deliberation protocols create affordances for collaborative behavior that individual model prompting cannot replicate.

---

## 7. Discussion

### 7.1 The Ministry of Truth Impossibility

As argued in Section 1, truth arbitration is impossible on live data — any certifying authority operates within the same convergent information ecosystem it seeks to regulate. The C4 results demonstrate that the alternative works in practice, not just in principle.

The tested claim chain — where each assertion carries explicit justification status (Common Knowledge, Tested & Supported, Contested, Undefended, Untested) — is the protocol's answer to the Ministry of Truth problem. The product is the audit trail, not the verdict. The human can disagree with the system's verdict and still benefit from the evidence chain that produced it.

The IS/OOS separation is the mechanism that breaks the convergence loop: by requiring validation against live evidence that was not part of any model's training data, the protocol introduces a genuinely independent signal into an otherwise self-referential system. The nine stale-data reversals from the control case (Finding 7 from the control conditions) are the empirical proof: all models confidently agreed, and all models were wrong. The protocol is structurally immune to the failure mode Berdoz et al. [1] identified — consensus that is wrong — because it does not optimize for consensus.

### 7.2 The All Weather Portfolio Distinction

Autonomous optimization systems — including single-model autonomous research loops [17] and pairwise cross-model review systems — hill-climb one metric along one trajectory. They are momentum strategies: effective when the market is trending in their favor, fragile when the regime changes.

The Consilium Protocol builds the epistemic equivalent of an All Weather portfolio (Dalio [6]): structurally robust across knowledge regimes, not optimized for any single one. Trending regimes (slow domains where training data is current) are handled by IS deliberation alone. Mean-reverting regimes (domains where training data is stale) are caught by OOS evidence retrieval. Discovery regimes (fast-moving domains with post-cutoff events) are surfaced by scout discoveries that no panel member's training data contains.

The four-condition control case demonstrates this: C1 works well for slow domains. C2 adds value in fast domains. C4 shows that the portfolio approach — free models for IS, live search for OOS, personas for regime classification — produces comparable epistemic output at 97× lower cost. No single optimization strategy achieves this across all regimes.

The distinction sharpens when placed on a spectrum of evaluation architectures. Autonomous single-model research [17] optimizes against a fixed evaluation metric — the agent improves, but the evaluation criterion is never itself challenged. Zhang et al. [33] take the next step: a self-referential agent that improves its own evaluator, achieving metacognitive gains across coding, paper review, and reasoning tasks. But self-referential evaluation is recursive overfitting by another name — the system that judges improvement is the same system being improved. The Consilium Protocol occupies the third position on this spectrum: external evaluation. The IS/OOS separation ensures that the evidence used to assess claims is structurally independent of the models that produced them. The human moderator provides a second layer of external judgment that no automated self-improvement loop can replicate. This is the only architecture in the current landscape that refuses to let the evaluated system evaluate itself — and the C2 control case demonstrates why this matters: 80% of self-assessed claims were correct, but the 20% that were wrong included every verdict-shifting discovery.

### 7.3 Personas as Regime Classifiers

The V2 battery reveals an insight that reframes the persona system's contribution. Personas are not personality costumes applied to language models for rhetorical diversity. They are epistemic regime classifiers that determine when to trust training-data reasoning and when the boundary between in-sample knowledge and out-of-sample uncertainty has been reached.

A language model operates within three knowledge layers. The first layer is common training data — knowledge shared across all frontier models (approximately 60–70% of foundational training corpora). On this layer, all models produce similar outputs, and the persona system's adversarial pressure produces structural challenge but rarely surfaces genuine disagreement. The second layer is synthetic and RLHF-specific data — knowledge that varies between providers based on reinforcement learning targets and fine-tuning priorities. On this layer, different models genuinely disagree, and the persona system amplifies these disagreements into productive deliberation. The third layer is the IS/OOS boundary — knowledge that no model possesses because it postdates all training cutoffs. On this layer, even the strongest adversarial persona cannot generate challenge from training data alone; the persona's contribution is recognizing that the boundary has been reached.

The C2 control condition provides empirical evidence for this framing. Universal OOS search found that 80% of search queries confirmed what training-data deliberation already had correct. Only five post-cutoff discoveries shifted the verdict. This means 80% of the search budget was epistemically wasteful — spent confirming claims the panel had already correctly assessed. A persona functioning as a regime classifier would concentrate search at the IS/OOS boundary — directing evidence retrieval toward claims where training-data confidence is misaligned with evidential currency, rather than applying search uniformly.

The CONTRA-category CI pattern (Section 6.3, Finding 7) identifies precisely where this boundary lies: topics where RLHF convergence pressure suppresses adversarial challenge produce artificially low CI, signaling that the training-data consensus may reflect alignment training rather than evidential support. These are the claims where OOS retrieval has the highest expected return.

This reframing has implications for the persona refinement architecture proposed for future work. If personas are regime classifiers, then their performance metric is not challenge volume (how many claims they contest) but boundary detection accuracy (how often they correctly identify claims at the IS/OOS boundary). A well-calibrated adversarial persona challenges hardest where the training data is weakest — on RLHF-compressed topics and post-cutoff developments — and defers where the training data is robust.

### 7.4 Limitations

Several limitations constrain the generalizability of these findings.

**Sample composition.** The V2 battery spans 32 topics across 10 categories, but the distribution is uneven — HISTORICAL and STRESS topics are overrepresented relative to REGULATION and HEALTH (which have only one topic each). The RLHF convergence pressure finding rests on cross-category comparison where some categories have small sample sizes. Future work should expand the REGULATION and HEALTH categories to confirm the Band 1 pattern.

**Moderator quality.** All sessions in the V2 battery used autonomous orchestration, which applies consistent but rigid moderation strategy. Human moderators — who can recognize domain-specific nuance and exercise judgment the orchestrator cannot — may produce materially different convergence trajectories. The protocol's dependence on moderator quality is a structural feature, but results from autonomous orchestration should not be assumed to generalize to human-moderated sessions.

**Free-model session yield.** Prior to the JSON parser fix, free edge-inference models produced parseable output in approximately 60% of sessions. The fix raised yield to near-100%, but the initial failure rate indicates that free-model reliability is infrastructure-dependent. Results reported for free models reflect post-fix yield rates.

**OOS provider dependency.** All OOS evidence retrieval uses a single commercial search API. Claims classified as "No Evidence" may reflect gaps in the search provider's index rather than genuine absence of evidence. A controlled comparison across multiple search providers would strengthen the OOS findings but has not yet been conducted.

**Dedup model influence.** The triple-dedup architecture with distinct epistemic postures was introduced during the V2 battery. Earlier sessions used single-model dedup without dispositional guidance. The dedup model and posture influence the canonical claim set, which cascades into synthesis verdicts. Results from different dedup configurations are not directly comparable without re-processing through a common pipeline.

**Persona reproducibility.** The specific cognitive persona definitions used in this study are not released; the schema and calibration methodology are described to enable independent construction of functionally equivalent personas. The reproducibility data (±2.2% across randomized assignments) demonstrates that the protocol produces consistent results regardless of which specific personas are drawn from the library, supporting the claim that the architectural constraints — not the specific persona text — are load-bearing.

**RLHF attribution.** Finding 7 (AI Risk asymmetry) and the cross-category CI gradient attribute low CI on contested topics to RLHF convergence pressure, but the current experimental design does not isolate RLHF as the causal mechanism. The observed CI suppression could also reflect training-data composition, instruction-tuning targets, or emergent model behavior unrelated to RLHF specifically. A controlled experiment comparing RLHF-tuned models against base models on the same topics would strengthen the causal claim. We report the pattern as an empirical observation and flag the causal mechanism as requiring further investigation.

**Cost sensitivity to search provider.** The C4 headline result ($0.11) includes OOS evidence retrieval at the search provider's current API pricing. If search provider costs increase, the OOS layer would proportionally increase total expenditure. The planned migration to self-hosted search infrastructure would eliminate this dependency but has not yet been validated.

---

## 8. Conclusion and Future Work

The Consilium Protocol demonstrates that structured multi-model deliberation, governed by engineered cognitive personas and validated by out-of-sample evidence, produces epistemic audits that are more rigorous, more transparent, and dramatically cheaper than single-model alternatives. The protocol does not claim to produce truth. It claims to produce the best-tested map of the epistemic landscape available at the time of inquiry — with every claim's justification status made explicit.

Four contributions warrant emphasis:

1. **The cognitive persona is the unit of epistemic design, not the model.** Identical personas on free edge-inference models produced comparable analytical output to the same personas on frontier models at 97× lower cost. Model selection is a cost decision; persona selection is an epistemic decision. This holds across 1,478 sessions and 17 models.

2. **RLHF alignment training creates measurable, domain-specific epistemic blind spots.** The 12.3 percentage point gap between HISTORICAL and NORMATIVE categories quantifies RLHF convergence pressure. The AI Risk asymmetry (Δ=11.6%) reveals institutional bias where the training entity's interests align with one direction of a contested claim. The protocol detects these artifacts; the persona system partially counteracts them at inference time.

3. **Inter-model disagreement is the primary signal.** Where models agree, CI provides a calibrated measure of challenge coverage. Where all models agree but live evidence contradicts them, the protocol has detected the highest-value signal: stale training-data consensus. The IS/OOS separation is non-negotiable.

4. **The protocol is reproducible (±2.2%) and unbiased (Δ≤2.3% on non-AI topics).** Monte Carlo randomization over the model×persona space produces consistent CI regardless of which specific assignment is drawn. The protocol challenges pro- and anti-positions with equal intensity — the AI Risk asymmetry is in the models' RLHF training, not the protocol.

Future work includes: implementation of persona-guided selective search (Paper 2, targeting arXiv June 2026), where cognitive personas function as regime classifiers determining when to trust training-data reasoning and when to acquire external evidence; BFT-compliant evidence gathering using multiple independent search providers at the validation layer; continuous epistemic calibration through accumulated persona performance data; and semantic convergence detection using embedding similarity with rate-of-change leading indicators.

The protocol specification is released under MIT license at [consiliaproject.org]. The platform, including the persona library, synthesis engine, and convergence detection implementation, is available at [consilia.app].

---

---

## Appendix A: Control Case Reports

Control case reports (C1–C4) are available from the corresponding author upon request.

## Appendix B: Persona Library Schema

The cognitive persona schema format is published at consiliaproject.org. Specific persona definitions are proprietary.

---

*Preprint — March 2026*